\pdfoutput=1
\documentclass[letterpaper, 10 pt, conference]{ieeeconf}  %
\IEEEoverridecommandlockouts
\usepackage[utf8]{inputenc}
\usepackage[T1]{fontenc}
\usepackage[noadjust]{cite}
\usepackage{url}
\usepackage{booktabs}
\usepackage{array}
\usepackage{amsmath,amssymb,amsfonts}
\usepackage{graphicx}
\usepackage{tikz}
\usetikzlibrary{arrows.meta}
\usepackage{pifont}
\usepackage{float}
\usepackage{hyperref}  %
\usepackage{subcaption}      %

\newcommand{\yes}{\ding{51}}
\newcommand{\no}{\ding{55}}
\newcommand{\pt}{$\sim$}

\newcommand{\RoboDyna}{\textbf{\textit{RoboDyna}}}

\title{\LARGE \bf RoboDyna: A Benchmark for Evaluating Robotic Manipulation under Physically Grounded Dynamics}

\author{Zhiyuan Li$^{1,2\dagger}$, Rui Heng Yang$^{1}$, Xuan Zhao$^{1}$, and Amir Rasouli$^{1,*}$%
\thanks{$^{1}$Huawei Technologies. $^{2}$University of Toronto.}%
\thanks{$^{*}$Corresponding author: {\tt\small amir.rasouli@huawei.com}}%
\thanks{$^{\dagger}$Work done during internship at Huawei Canada.}%
}

\begin{document}
\bstctlcite{IEEEabrv}

\maketitle
\thispagestyle{empty}
\pagestyle{empty}
\begin{abstract}
Object manipulation in the dynamic real world requires understanding how the environment functions and evolves, and acting accordingly. Existing benchmarks, however, do not adequately address these needs. Physics-oriented benchmarks evaluate predictive fidelity but not closed-loop action. Dynamic robotic manipulation benchmarks often use simplified, physically implausible object motions, limiting evaluation of reasoning and action under realistic dynamics. Evaluation is also typically synchronized, freezing the environment during inference and making computation time irrelevant to task outcomes. 
To this end, we propose \RoboDyna{}, a benchmark for evaluating the physical understanding and action capabilities of manipulation policies in dynamic scenes. \RoboDyna{} comprises 20 conceptual dynamic tasks spanning five capability categories, with simple objects and minimal scene clutter, under a two-factor difficulty design, and 10  household tasks that instantiate the same capabilities in visually and semantically richer simulated real-world environments. \RoboDyna{} provides 3{,}500 expert demonstrations, paired synchronous and asynchronous evaluation protocols, metrics for task success and partial completion, and interactive tools for human baseline evaluation.
Evaluating vision-language-action models and a world-action model across 5.4K episodes, we find that all policies achieve success rates below 20\%, revealing substantial limitations in dynamic manipulation. The world-action model captures simple physical dynamics but struggles with complex interactions, and its prediction fidelity does not consistently align with task success. Sync--async comparisons show that inference latency flips 5--12\% of episode outcomes while changing average success rates by at most 3\%. Human baselines substantially outperform all policies under matched conditions.  
The benchmark and complete results are available at \url{https://robodyna.github.io/robodyna/}.

\end{abstract}

\section{Introduction}

Recent robotic manipulation policies demonstrate impressive capabilities in static scenes~\cite{black2024pi0,pi2025pi05,liu2023libero,nasiriany2024robocasa}, but their physical competence in dynamic environments remains underexplored. Completing a task in a dynamic environment introduces two challenges~\cite{fang2026domino,xie2026dynamicvla}. First, selecting the correct action may require integrating multiple observations to infer object motion and anticipate future states. Second, the correct action must be executed within a critical time window, as scene evolution can render an otherwise valid action ineffective~\cite{chen2026reflex,black2025rtc}. %

Evaluating such physical competence requires both anticipating how the scene will evolve and acting accordingly. However, current benchmarks do not jointly evaluate both. Predictive world-model benchmarks~\cite{nvidia2025cosmos,liao2025genie} primarily evaluate physical consistency and the fidelity of predicted futures, with limited emphasis on closed-loop policy performance under realistic execution timing. Robotics benchmarks either contain no independent scene dynamics~\cite{gu2023maniskill2,liu2023libero,nasiriany2024robocasa} or animate targets with scripted motion while pausing the world during inference~\cite{fang2026domino}.
Evaluating under the latter regime has several consequences. First, \textit{physically implausible dynamics limit transfer to real-world physics}. Second, \textit{limited dynamic complexity restricts the range of physical capabilities tested}. Third, \textit{synchronous evaluation makes inference cost-free}: the environment is frozen during computation, whether a policy takes seconds per action chunk~\cite{feng2025vidarc} or runs at tens of hertz~\cite{black2024pi0}.

\begin{figure*}[t]
  \centering
  \includegraphics[width=0.85\textwidth]{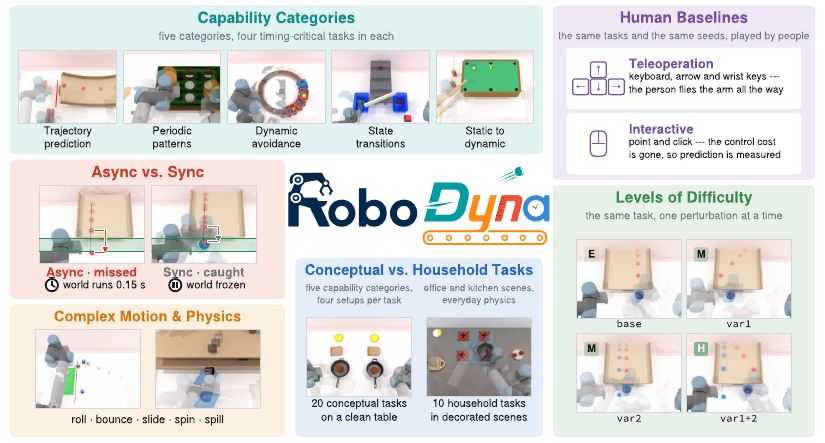}
  \caption{\textbf{\RoboDyna{} at a glance.} 20 conceptual tasks in five
  capability categories and 10 household tasks with physics-governed motion;
  four difficulty settings per conceptual task; paired sync/async protocols;
  and a human study on the same seeds.}
  \label{fig:teaser}
\end{figure*}

To this end, we propose \RoboDyna{}, a \textit{physics-governed, latency-aware benchmark} for dual-arm manipulation in dynamic environments (see
Fig.~\ref{fig:teaser}). \RoboDyna{} challenges policies to anticipate scene evolution, accounts for inference latency, and jointly evaluates physical understanding and action capabilities.
The proposed benchmark consists of \textit{20 conceptual tasks}, divided into five physical-understanding categories with varying difficulty levels to measure core policy competencies. An additional \textit{10 complex household tasks} are designed to evaluate transfer to realistic scenes. For evaluation, we use \textit{mixed metrics} to record full and partial success under \textit{synchronous and asynchronous} regimes. Moreover, our benchmark contains 3{,}500 expert demonstrations and tools for collecting human baseline performance under teleoperation and scene-interactive control.
In summary our contributions are as follows:  (1) we propose \RoboDyna{}, a benchmark suite of
\textbf{physics-governed dynamic tasks}, which consists of 20 conceptual tasks in five capability categories with four per-task difficulty designs and additional 10 household tasks; (2) We define a paired synchronous and asynchronous evaluation protocol to isolate the effect of inference latency on task outcome and mixed success and partial-success metrics;  (3) We conduct extensive evaluation of two state-of-the-art VLAs and one WAM on over 5400 episodes to characterize their physical competence under different dynamic conditions; (4) We report  on 12-participant human baseline performance on proposed tasks using different control regimes.

\section{Related Work}
\label{sec:related}

\subsection{Robotic Manipulation Policies}
Learning-based robotic action policies, such as vision-language-action models (VLAs)~\cite{pi2025pi05,zheng2025xvla,black2024pi0,kim2024openvla} and world-action-models (WAMs) \cite{liao2025genie,kim2026cosmospolicy,zheng2026motionwam,yuan2026fastwam}, map images, language, and robot states to actions, with recent systems spanning flow-matching, autoregressive, and video-prediction-based designs. Standard evaluations primarily measure task success, often using a synchronous simulator or fixed action clock. While useful for comparing grounding and motor competence, such evaluations do not reveal whether a policy anticipates rolling, bouncing, or contact-induced motion, nor whether its computation finishes in time to act on such dynamics. We evaluate two VLAs, $\pi_{0.5}$~\cite{pi2025pi05} and X-VLA~\cite{zheng2025xvla}, and one WAM, FastWAM~\cite{yuan2026fastwam}, to compare their performance on dynamic manipulation tasks.

\subsection{Predicting and Understanding Physics}
Video and world-model benchmarks evaluate future prediction using image or state fidelity, perceptual similarity, or physical-plausibility scores~\cite{nvidia2025cosmos,liao2025genie,feng2025vidarc}. These protocols assess whether a model can forecast how a scene will evolve, but not whether it can act successfully in that evolving scene. A visually accurate forecast can coexist with an action that is too late, misdirected, or dynamically infeasible, for example, causing unintended collisions between the robot and scene objects. Closed-loop efforts, such as World-in-World~\cite{zhang2025worldinworld} incorporate actions into prediction, yet do not evaluate robotic manipulation under continuous physical dynamics and measured inference latency. 
\RoboDyna{} instead evaluates physical reasoning through action, testing whether policies can successfully manipulate objects under gravity, friction, contact, and continuous scene evolution.

\subsection{Robotic Manipulation Benchmarks}
\noindent\textbf{Dynamic design.} General-purpose suites, such as RoboDojo
~\cite{chen2026robodojo}, RoboCasa~\cite{nasiriany2024robocasa}, and LIBERO
~\cite{liu2023libero} evaluate complex manipulation, robustness, and
long-horizon composition. Their canonical scenes are largely static, with
scene changes driven primarily by the robot's actions rather than independent dynamics. Emerging benchmarks with object motion~\cite{fang2026domino} animate the targets of otherwise static tasks with scripted motion. The resulting dynamics are not physically plausible: objects translate sideways or drift randomly through the scene, unconstrained by gravity or friction. \RoboDyna{} instead builds its tasks around physically grounded dynamics. Five capability categories cover complex motion in which objects roll, bounce, slide, and spin under gravity and friction (Section~\ref{sec:benchmark}). Across these settings, we probe how policy performance degrades with increasing dynamic complexity. As Table~\ref{tab:related} shows, \RoboDyna{} covers a broader range of physics-governed dynamic phenomena than existing manipulation benchmarks.

\begin{figure*}[t]
  \centering
  \includegraphics[width=\textwidth]{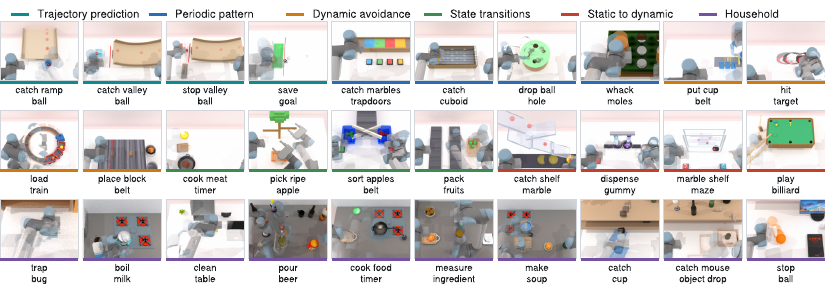}
  \caption{\textbf{The 30 \RoboDyna{} tasks.} Rows one and two are the 20 conceptual tasks in category order;
  row three is the 10 household scenes. }
  \label{fig:gallery-sub}\vspace{-0.4cm}
\end{figure*}

\begin{table}[t]
\centering
\caption{Comparison along the three benchmark-design axes. \yes=yes,
\no=no, and \pt=partial.}
\label{tab:related}
\scriptsize
\setlength{\tabcolsep}{3pt}
\renewcommand{\arraystretch}{1.15}
\resizebox{\columnwidth}{!}{%
\begin{tabular}{@{}lccc@{}}
\toprule
\textbf{Benchmark} & \shortstack{\textbf{Dynamic capability}\\\textbf{taxonomy}} & \shortstack{\textbf{Physics-grounded}\\\textbf{dynamics}} & \shortstack{\textbf{Paired}\\\textbf{Sync./Async.}} \\
\midrule
Static-scene benchmarks & \no & \no & \no \\
DOMINO~\cite{fang2026domino} & \pt & \no & \no \\
DOM~\cite{xie2026dynamicvla} & \pt & \pt & \pt \\
ReflexBench~\cite{chen2026reflex} & \no & \pt & \pt \\
\textbf{\RoboDyna{} (ours)} & \yes & \yes & \yes \\
\bottomrule
\end{tabular}
}\vspace{-0.4cm}
\end{table}

\noindent\textbf{Inference Latency.} 
Standard robotic manipulation benchmarks typically freeze the world during policy inference. Under this synchronous convention, a slow policy receives the same state as a fast one, removing the need to account for scene evolution during inference. Concurrent asynchronous
protocols~\cite{xie2026dynamicvla,chen2026reflex} relax this assumption, but do not provide a paired synchronous/asynchronous evaluation on identical physics-grounded episodes. \RoboDyna{} fills this gap by replaying identical seeds under synchronous and asynchronous protocols, where the latter allows the scene to evolve during policy inference. This paired evaluation isolates the effect of inference latency on task performance.

\section{The \RoboDyna{} Benchmark}
\label{sec:benchmark}

\subsection{Problem Formulation}
\label{sec:formulation}
Each \RoboDyna{} task is a language-conditioned dual-arm manipulation problem in SAPIEN~\cite{xiang2020sapien}, with a scene that evolves continuously under physical dynamics. Formally, a policy $\pi_\theta(a_{t+n}\mid o_{t:t+n},\ell)$ maps a history of observations $o_{1:n}$ (RGB streams and proprioception) and a language instruction $\ell$ to a dual-arm action chunk $a_{t+n}\in\mathbb{R}^{H\times14}$, where $H$ is the length of the action chunk and each dual-arm action has 14 dimensions. The scene state $s(t)$ evolves continuously and independently of policy inference. Each episode has a task-specific success predicate $g(s_T)\in{0,1}$ evaluated on the simulator state. Let $\Delta_n$ denote the measured wall-clock inference time for action chunk $a_n$. During this interval, the scene may evolve such that an action selected from the current observation is no longer appropriate at execution time.

We model this timing effect with an execution operator $\mathcal{E}^{\delta}$:
the action chunk $a_n$, computed from observations available at time $t_n$, is
executed after $\delta\Delta_n$ seconds of scene evolution. Setting $\delta=0$
yields synchronous evaluation, whereas $\delta=1$ yields asynchronous
evaluation. The benchmark objective is
\begin{equation}
J(\pi_\theta;\delta)
=
\mathbb{E}_{\tau\sim\mathcal{D}}
\left[
g\left(s_T^{\pi_\theta,\delta}\right)
\right],
\label{eq:objective}
\end{equation}
where $\mathcal{D}$ is the distribution of task configurations and episode
seeds, and $s_T^{\pi_\theta,\delta}$ denotes the terminal state obtained by
executing $\pi_\theta$ under $\mathcal{E}^{\delta}$. Thus,
$J(\pi_\theta;0)$ and $J(\pi_\theta;1)$ give paired synchronous and
asynchronous success rates on the same task instances.

\subsection{Dynamic Task Design}
\label{sec:dynamic-design}
\RoboDyna{} is designed to systematically probe physical competence across diverse dynamic phenomena. We organize these dynamics into five capability categories. \emph{Trajectory prediction} requires anticipating the future position of objects undergoing continuous motion, including motion redirected by contact or collision. \emph{Periodic-pattern understanding} requires identifying recurring motion patterns and timing actions to transient opportunity windows. \emph{Dynamic avoidance} requires planning robot motion around independently moving objects and obstacles. \emph{State-transition estimation} requires tracking changes not captured by object pose alone, such as appearance or object availability. Finally, \emph{static-to-dynamic understanding} requires anticipating motion induced by the robot's own actions from an initially static scene, such as through pushing, striking, or releasing an object.

In \RoboDyna{}, task-critical object motion arises from simulated rigid-body dynamics--- gravity, friction, collision, and contact --- rather than scripted target trajectories. A subset of scene processes (e.g. the train moving along a track) are driven by task-defined kinematic or temporal update rules. We carefully configure object properties, physical parameters, and interaction mechanisms so that the resulting behaviors remain physically plausible and consistent with their real-world counterparts. Compared to existing benchmarks, this grounding reduces the gap introduced by arbitrary scripted dynamics and makes benchmark performance more indicative of physical competence relevant to real-world manipulation.

Within each dynamic setting, we randomize task-relevant initial conditions and physical parameters, including object positions, motion directions and velocities, interaction geometry, and task-specific dynamic properties. As a result, repeated episodes can produce different trajectories, contact events, and event timings while preserving the underlying capability being evaluated. This variability reduces reliance on memorizing fixed trajectories and requires policies to reason about the dynamics instantiated in each episode.

\subsection{Tasks and Difficulty Variations}
\label{sec:task-variations}

\begin{figure}[!t]
  \centering
  \begin{minipage}[t]{0.24\columnwidth}
    \centering
    \includegraphics[width=\linewidth,trim=140 80 240 210,clip]{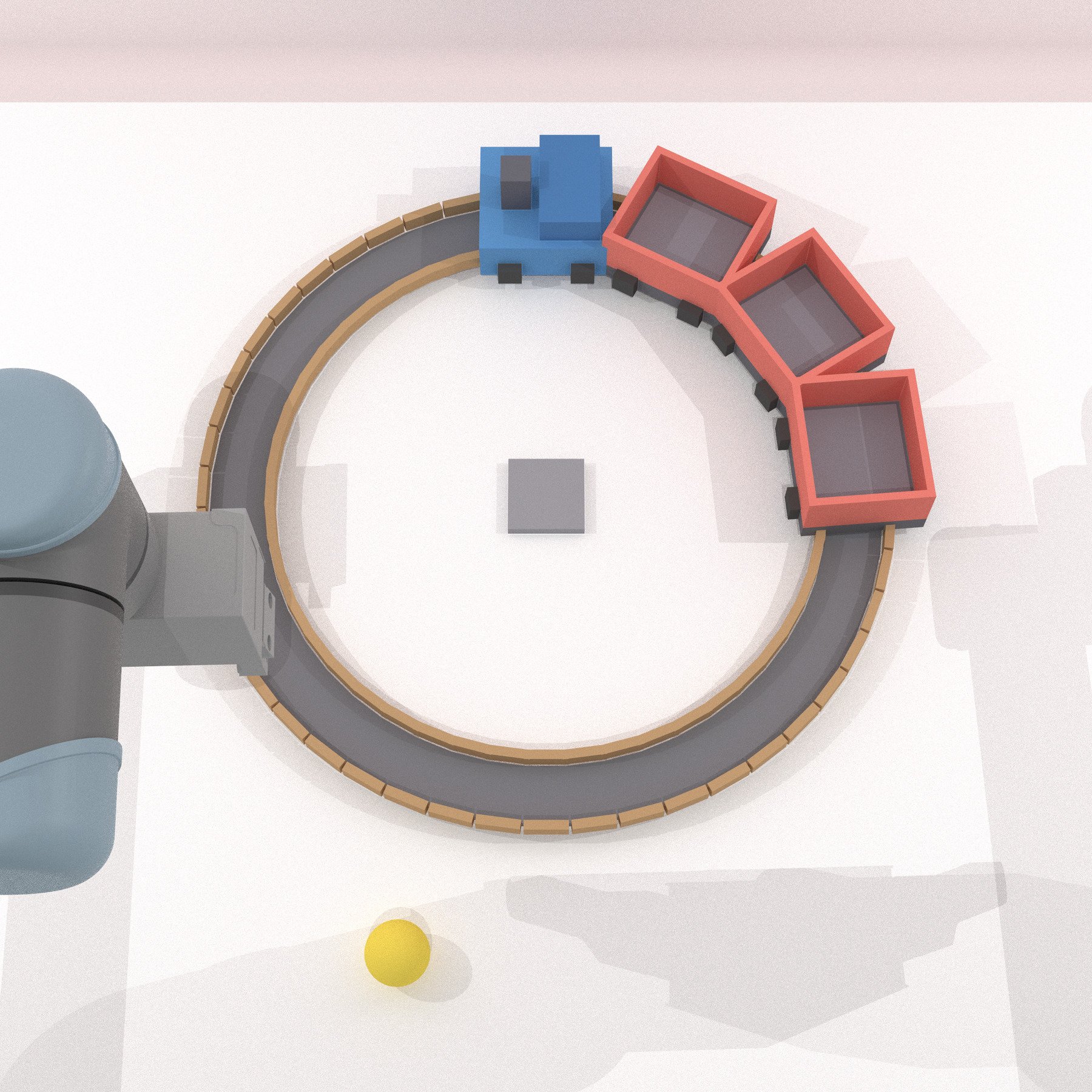}\\[-1pt]
    {\scriptsize (a) \texttt{base}\\any wagon}
  \end{minipage}
  \begin{minipage}[t]{0.24\columnwidth}
    \centering
    \includegraphics[width=\linewidth,trim=140 80 240 210,clip]{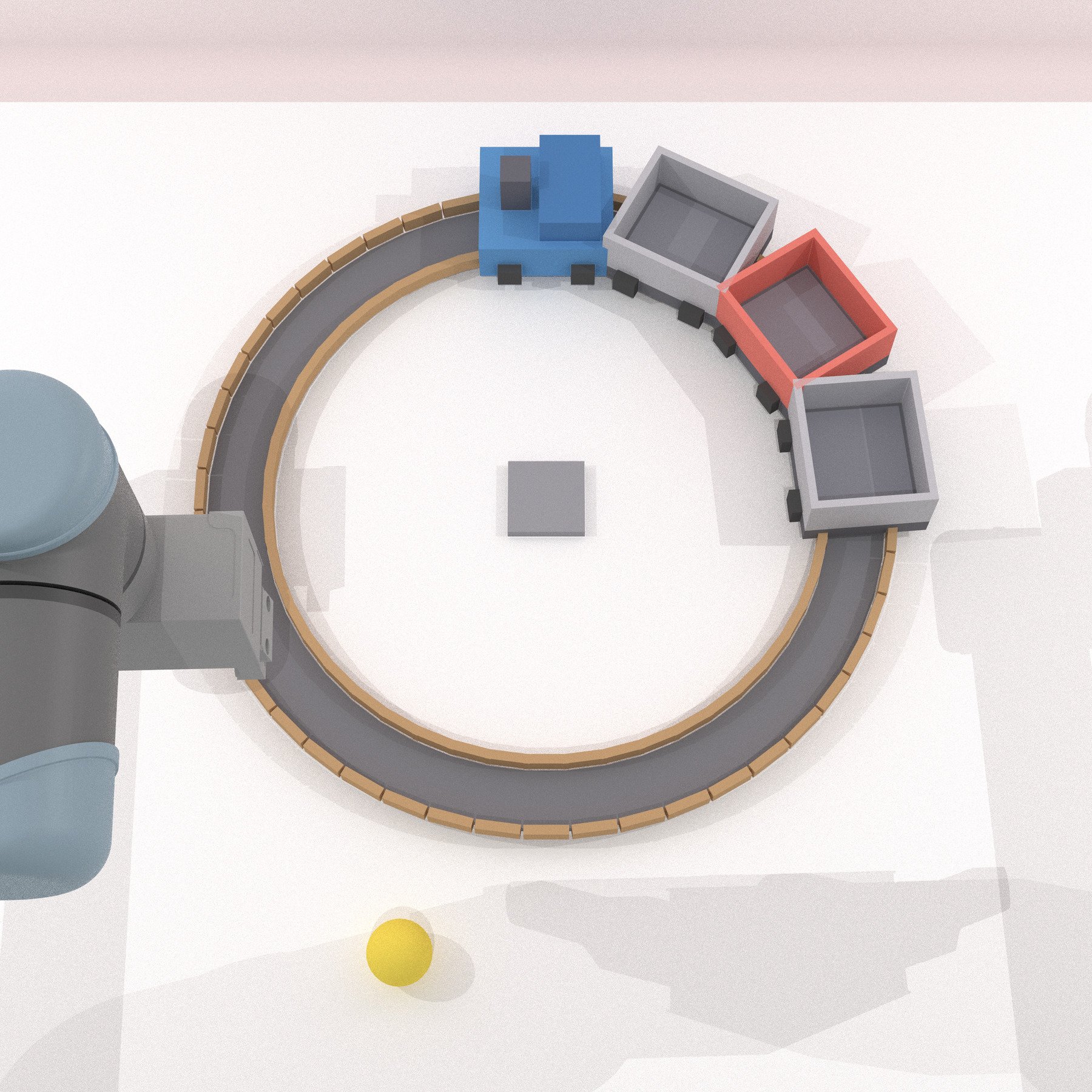}\\[-1pt]
    {\scriptsize (b) \texttt{var1}\\red wagon only}
  \end{minipage}
  \begin{minipage}[t]{0.24\columnwidth}
    \centering
    \includegraphics[width=\linewidth,trim=140 80 240 210,clip]{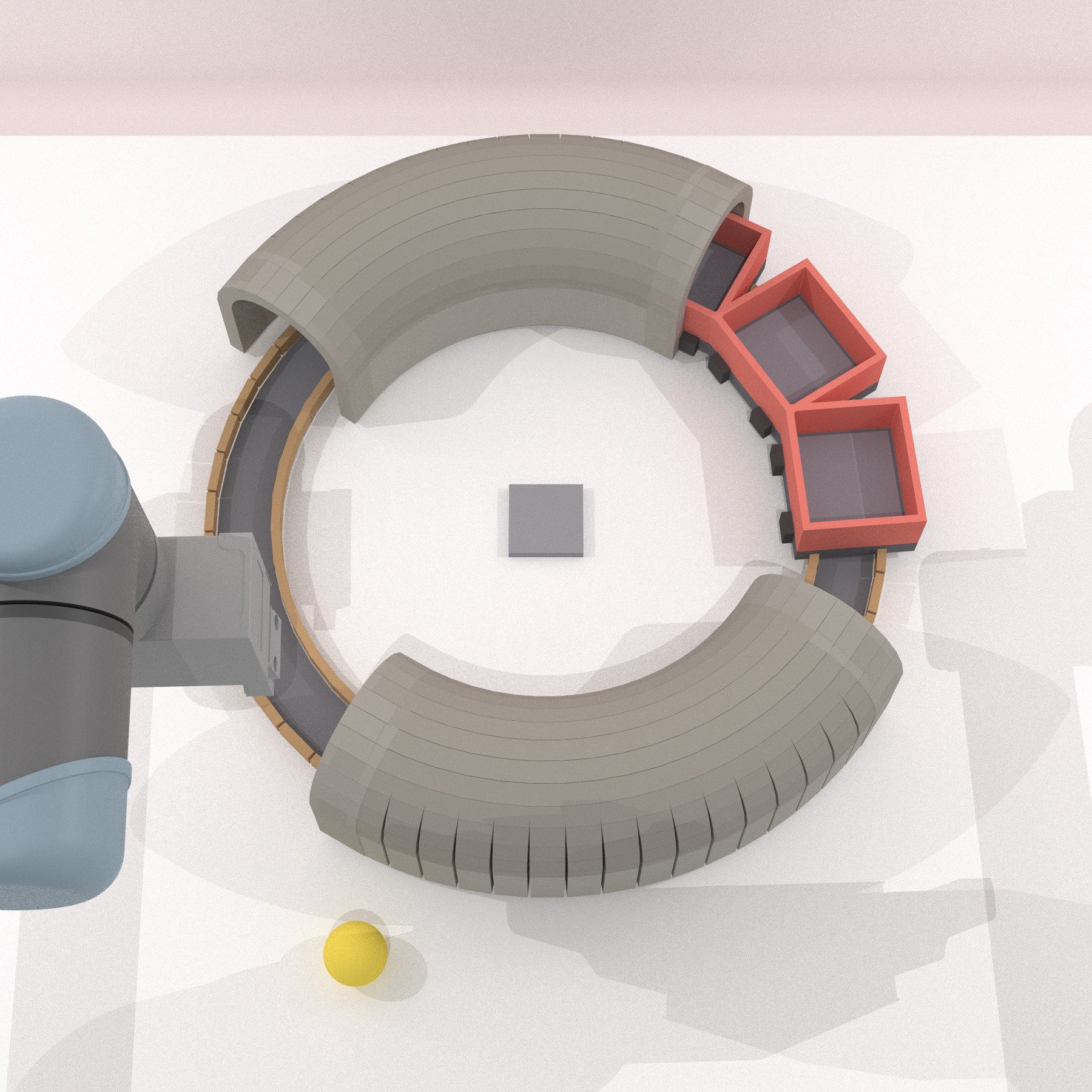}\\[-1pt]
    {\scriptsize (c) \texttt{var2}\\tunnels}
  \end{minipage}
  \begin{minipage}[t]{0.24\columnwidth}
    \centering
    \includegraphics[width=\linewidth,trim=140 80 240 210,clip]{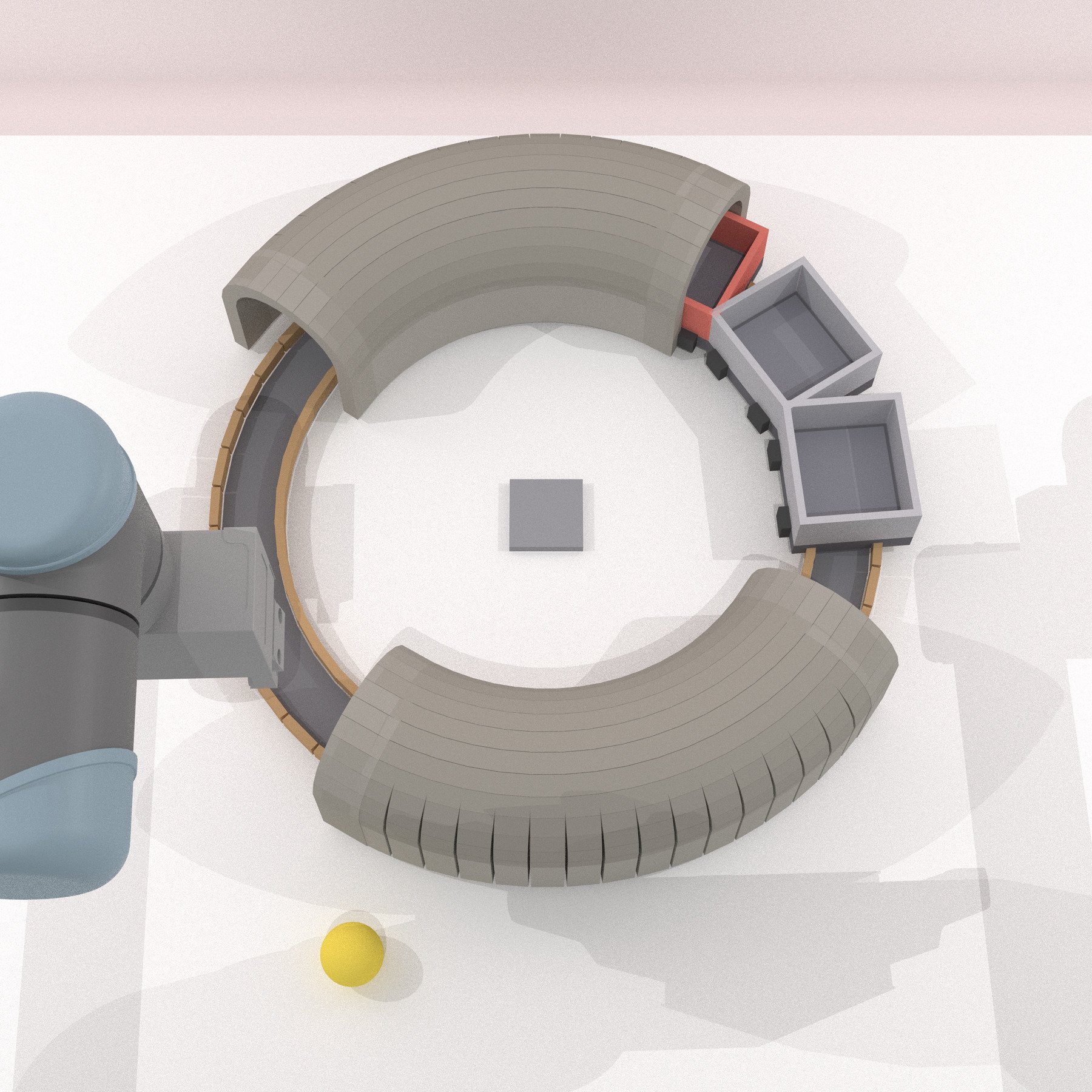}\\[-1pt]
    {\scriptsize (d) \texttt{var1+2}\\both}
  \end{minipage}
  \caption{\textbf{Difficulty variations in load train.} The robot places the
  yellow ball into a passing red wagon. Gray wagons are invalid targets. } \vspace{-0.3cm}
  \label{fig:load-train-variations}
\end{figure}

\noindent\textbf{Conceptual Tasks.} 
To instantiate and isolate the five dynamic capabilities, the 20 \emph{conceptual} tasks use minimal scenes containing only task-relevant objects, such as geometric primitives and common objects. Each conceptual task has four conditions: \texttt{base} (easy); \texttt{var1} and \texttt{var2} (medium), each introducing one task-specific variation; and \texttt{var1+2} (hard), which composes both variations and is withheld from the training demonstrations. Base conditions contain the simplest dynamics for each task, for example an object moving along a straight trajectory. Variations increase difficulty by altering the underlying dynamics or adding interaction constraints. These include contact-induced motion such as bouncing, changes in physical properties such as friction, patterned motion of distractors, obstacles that introduce collision risk or partial observability, and more restrictive success conditions. For instance, in the \emph{load train} task (Fig.~\ref{fig:load-train-variations}), the robot must place a ball into a wagon of a train moving around a circular track. In \texttt{base}, any open wagon is a valid target; \texttt{var1} restricts the target to the red wagon; \texttt{var2} adds tunnels that occlude the train and block placement along the covered track; \texttt{var1+2} combines both constraints. Across tasks, \texttt{var1} and \texttt{var2} are designed as independent modifications of the base condition, such that \texttt{var1+2} tests their composition rather than introducing a new source of difficulty.

\begin{figure}[!t]
\centering
\begin{minipage}[t]{0.49\columnwidth}
\centering
\begin{tikzpicture}
\node[anchor=south west,inner sep=0] (im) at (0,0) {\includegraphics[width=\linewidth,height=.79\linewidth,trim=25 83 291 185,clip]{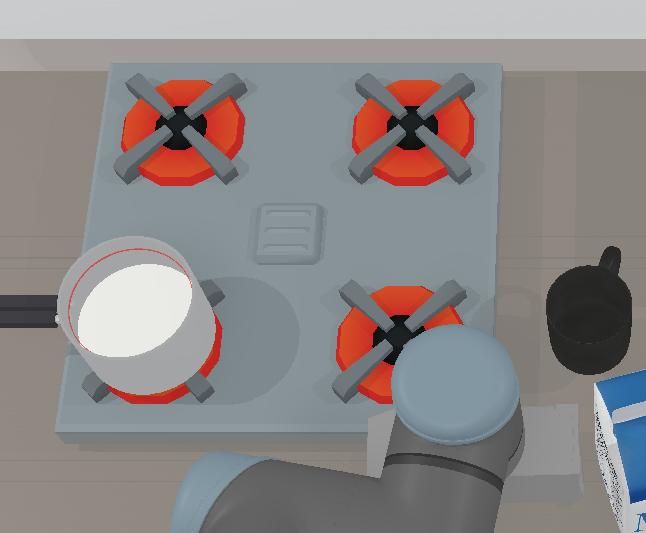}};
\begin{scope}[x={(im.south east)},y={(im.north west)},every node/.style={font=\sffamily\fontsize{6}{7}\selectfont,fill=white,fill opacity=.94,text opacity=1,inner sep=1.3pt,align=center},every path/.style={line width=.65pt}]
\node (stop) at (.73,.89) {Stop at mark};
\draw[red!70!black,-{Stealth[length=1.2mm]}] (stop.south west) -- (.35,.75);
\node (rise) at (.73,.20) {Milk rises};
\draw[orange!85!black,-{Stealth[length=1.2mm]}] (rise.north west) -- (.39,.47);
\end{scope}
\end{tikzpicture}
\par\smallskip
{\footnotesize (a) Boil milk}
\end{minipage}\hfill
\begin{minipage}[t]{0.49\columnwidth}
\centering
\begin{tikzpicture}
\node[anchor=south west,inner sep=0] (im) at (0,0) {\includegraphics[width=\linewidth,height=.79\linewidth,trim=180 43 0 10,clip]{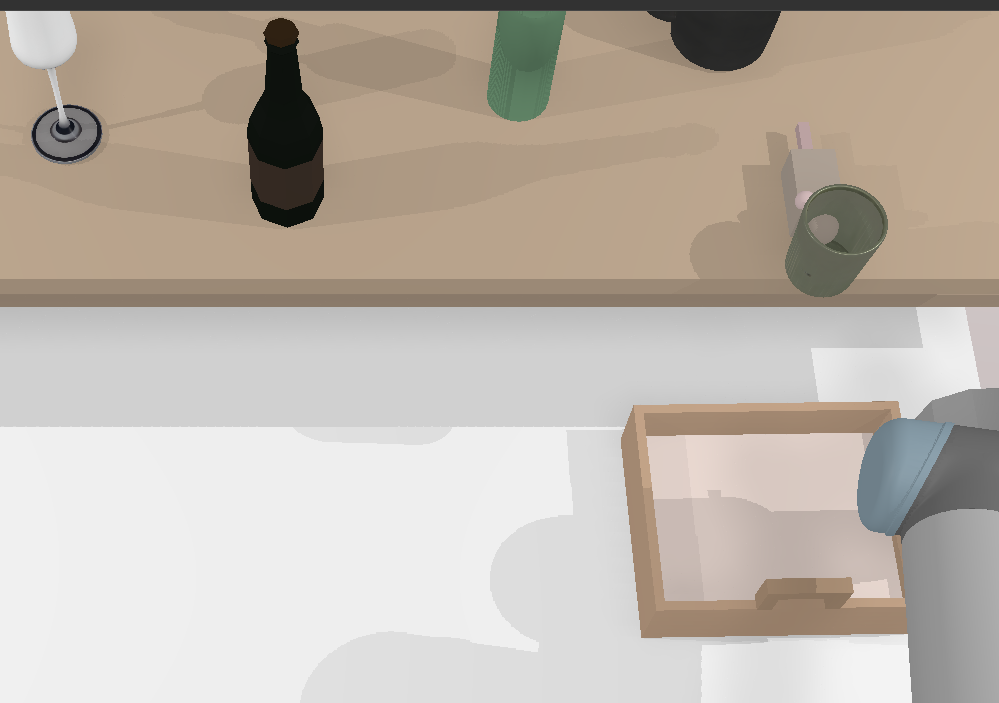}};
\begin{scope}[x={(im.south east)},y={(im.north west)},every node/.style={font=\sffamily\fontsize{6}{7}\selectfont,fill=white,fill opacity=.94,text opacity=1,inner sep=1.3pt,align=center},every path/.style={line width=.65pt}]
\node (others) at (.29,.45) {Other possible\\push targets};
\draw[yellow!90!orange,line width=.9pt,-{Stealth[length=1.3mm]}] (others.north west) -- (.13,.77);
\draw[yellow!90!orange,line width=.9pt,-{Stealth[length=1.3mm]}] (others.north east) -- (.41,.88);
\node (mouse) at (.76,.94) {Mouse pushes};
\draw[orange!85!black,-{Stealth[length=1.2mm]}] (mouse.south) -- (.77,.75);
\draw[orange!85!black,dashed,-{Stealth[length=1.3mm]}] (.80,.57) .. controls (.81,.47) and (.78,.40) .. (.75,.30);
\node (catch) at (.30,.27) {Catch below};
\draw[black!75,-{Stealth[length=1.2mm]}] (catch.east) -- (.68,.24);
\end{scope}
\end{tikzpicture}
\par\smallskip
{\footnotesize (b) Mouse-induced object drop}
\end{minipage}
\caption{\textbf{Household dynamics.} (a) Milk rises during heating; the robot
must switch off at the marked level before overflow. (b) Several shelf objects are potential push targets. The robot must identify
the object the mouse dislodges and position
the basket below.} \vspace{-0.3cm}
\label{fig:household-dynamics}
\end{figure}

\noindent\textbf{Household tasks.} The household tasks instantiate the same dynamic capabilities in more visually and semantically complex office and kitchen scenes. Unlike the minimal conceptual scenes, household environments contain task-irrelevant objects, furniture, and visual clutter. They additionally introduce everyday physical phenomena beyond rigid-body motion, including fluid levels, overflow, dispensing, and state transitions. For example, \emph{boil milk} requires stopping heating before overflow, instantiating the same \emph{state-transition estimation} capability as the conceptual tasks through a new physical phenomenon---a continuously rising liquid level. In \emph{catch mouse object drop}, the policy must identify which object a moving mouse will dislodge and anticipate its subsequent falling trajectory, extending \emph{trajectory prediction} from single-object motion in the conceptual tasks to interactions involving multiple objects. Together, these tasks evaluate the same underlying dynamic capabilities under richer perceptual, temporal, and physical constraints than the capability-isolating conceptual tasks. Fig.~\ref{fig:household-dynamics} highlights the rising milk level and the mouse-induced object fall.

\subsection{Human Performance Collection}
Our benchmark includes interactive tools for collecting human demonstrations and evaluating human performance. For this purpose we use two regimes: direct robot teleoperation and scene-interactive control.
\subsubsection{Teleoperation} In this mode, the user directly controls the robot arm in the scene and takes charge of all actions, including grasping, pushing, and releasing. The end-effector's (EE's) $x$ and $y$ coordinates are controlled using the arrow keys, while its $z$ coordinate is controlled using the Q and E keys. Grippers opening and closing is controlled with the space bar and two additional key sets rotate wrist  and tilt the arm left and right. Since the controls command end-effector (EE) Cartesian targets rather than individual joints, we bound the EE z coordinate to avoid arm and wrist configurations from which recovery is difficult. In the case of an unrecoverable pose, a reset button can bring the arm to its original pose. In this mode, the user can switch between camera views that are available to the policy, including head and gripper cameras.  
\subsubsection{Scene-interactive} This mode removes the burden of low-level control so the participants can focus on task reasoning. Hence, the arms are removed from the scene and participants act directly on objects. Users control objects through mouse clicks, which specify placement locations on the $xy$ plane, and arrow keys, which select between buttons in scenes with two buttons. In this mode, the user only has a single fixed view of the head camera.  

\section{Dataset and Evaluation}
\label{sec:dataset-human}
\subsection{Implementation}
We implement \RoboDyna{} in SAPIEN~3.0.3 with a fixed dual-UR5 embodiment. Policies observe synchronized RGB streams from one head-mounted and two wrist-mounted Intel RealSense D435 cameras at $320\times240$ resolution, together with the joint and gripper states of both arms. The simulator runs at $250\,\mathrm{Hz}$ (a $4\,\mathrm{ms}$ physics timestep), and each action is held for 15 simulation steps, resulting in a $60\,\mathrm{ms}$ control interval ($16.7\,\mathrm{Hz}$). This matches the action cadence used to collect the expert demonstrations.

\subsection{Expert Data}
Demonstrations are generated by task-specific expert controllers that use simulator states and known dynamics to select end-effector targets and schedule grasps, interceptions, and releases. cuRobo~\cite{sundaralingam2023curobo} converts these targets into joint trajectories through inverse kinematics and motion planning under the configured collision constraints. SAPIEN's PhysX backend~\cite{xiang2020sapien} simulates rigid-body dynamics and contacts during execution, while prescribed motions are implemented through task-specific dynamics calculations or kinematic updates. For example, in \emph{Catch Ramp Ball}, we determine the ball's position and velocity at the ramp exit and estimate its subsequent ballistic trajectory to place the cup below its flight path, beyond the ramp. The ball is released into PhysX simulation at the ramp exit, and task completion is verified by checking whether it lands inside the cup. This procedure combines spatial and temporal planning with simulation-based verification to generate demonstrations that achieve the task through the intended physical interaction.

We provide 3{,}500 expert demonstrations across 70 task configurations. For each of the 20 conceptual tasks, we collect 50 demonstrations for each of the \texttt{base}, \texttt{var1}, and \texttt{var2} configurations, while the 20 \texttt{var1+2} configurations are withheld from the demonstration set for compositional evaluation. We additionally collect 50 demonstrations for each of the 10 household tasks. Across demonstrations, task-relevant initial conditions, including object positions, velocities, motion parameters, and target goals, are randomized to reduce reliance on fixed scene configurations or trajectories. Language instructions are instantiated according to the task and configuration.

\noindent \textbf{Training.} The three policies, namely $\pi_{0.5}$~\cite{pi2025pi05}, X-VLA~\cite{zheng2025xvla}, and FastWAM~\cite{yuan2026fastwam} are fine-tuned on the full set of 3{,}500 demonstrations and evaluated from a single checkpoint each ($\pi_{0.5}$ and X-VLA at 30k gradient steps, and FastWAM at 13k). The policies differ in their execution horizons before receiving a new observation: $\pi_{0.5}$ executes all 50 predicted actions, X-VLA all 30, and FastWAM the first 24 of 32, following their respective official implementations. Because inference latency and execution horizon differ across policies, asynchronous performance reflects both action capability and execution timing.

\subsection{Evaluation Protocols}
\label{sec:protocol}
\subsubsection*{Synchronous vs asynchronous}
We evaluate each policy under paired synchronous and asynchronous protocols using identical episode seeds. Under \emph{synchronous} evaluation, simulation is paused during policy inference. Under \emph{asynchronous} evaluation, the wall-clock duration of each forward pass is measured, and the scene continues to evolve for the corresponding simulated duration before the predicted action chunk is executed. This protocol models blocking inference and does not overlap inference with execution of the preceding chunk, as in real-time action chunking~\cite{black2025rtc}.

Each task--configuration--protocol combination is evaluated on the same ten held-out seeds across policies, yielding 900 episodes per policy per protocol and 5{,}400 evaluation episodes in total. Episode length is capped per task at 1.2 times the longest expert episode for that task. The human study uses the same episode seeds, enabling comparison on matched task instances.

\subsubsection*{Metrics}
We report both binary task success and a task-specific \emph{progress score} (PS) to quantify partial completion when full success is sparse. For each task, a base progress value $b\in[0,1]$ is determined by observable completion milestones, while $m$ counts predefined penalty events. We compute
\begin{equation}
\mathrm{PS}=b\cdot2^{-m},
\end{equation}
such that progress remains measurable even when binary success is zero. We use a fixed half-credit penalty: each predefined error event halves the progress score, which penalizes repeated errors while retaining credit for partial completion. Milestones and penalties are defined separately for each task according to its objective. For example, in \emph{Hit Target}, $b\in\{0,0.25,0.5,0.75,1\}$, with higher values assigned as the dart lands closer to the center of the target; each hit on a distractor increments $m$, halving the resulting score. Household tasks typically define milestones over task-relevant physical quantities. For example, in \emph{Pour Beer}, fill-level ranges of $55$--$65\%$, $65$--$75\%$, and $75$--$85\%$ receive progress values of $0.25$, $0.5$, and $0.75$, respectively, while overflow receives zero progress. %

\subsubsection*{Video prediction fidelity}
Following world-model evaluation~\cite{huang2025vid2world}, we measure prediction fidelity using PSNR and SSIM between predicted and recorded future head-camera images, with higher values indicating greater similarity. We average PSNR and SSIM over six future horizons (4 to 24 control steps, 0.24--1.44\,s), excluding the conditioning frame, weighting tasks equally.

\begin{table*}[t]
  \centering
  \setlength{\tabcolsep}{1pt}
  \footnotesize
  \renewcommand{\arraystretch}{1.12}
  \caption{Success rate and progress score (SR/PS) under Sync and Async evaluation. Mean inference time is shown below each policy. Bold values in each column indicate the best performance.}
  \label{tab:results-sub}
  \resizebox{\textwidth}{!}{%
  \begin{tabular}{@{}llcc@{\hspace{3pt}}cc@{\hspace{3pt}}cc@{\hspace{3pt}}cc@{\hspace{3pt}}cc@{\hspace{3pt}}cc@{\hspace{3pt}}c@{\hspace{3pt}}c@{}}
    \toprule
    & & \multicolumn{2}{c}{\shortstack{Traj.\\pred.}} & \multicolumn{2}{c}{\shortstack{Periodic\\pattern}} & \multicolumn{2}{c}{\shortstack{Dyn.\\avoid.}} & \multicolumn{2}{c}{\shortstack{State\\trans.}} & \multicolumn{2}{c}{\shortstack{Static to\\dyn.}} & \multicolumn{2}{c}{\shortstack{Conceptual\\Avg.}} & \multicolumn{1}{c}{\shortstack{Household\\Avg.}} & \multicolumn{1}{c}{\shortstack{Overall\\Avg.}} \\
    \cmidrule(lr){3-4} \cmidrule(lr){5-6} \cmidrule(lr){7-8} \cmidrule(lr){9-10} \cmidrule(lr){11-12} \cmidrule(lr){13-14}
    Policy & Protocol & Easy & Medium & Easy & Medium & Easy & Medium & Easy & Medium & Easy & Medium & Easy & Medium &  &  \\
    \midrule
    $\pi_{0.5}$ & Sync & \textbf{0.23}/\textbf{0.26} & \textbf{0.23}/\textbf{0.28} & \textbf{0.35}/\textbf{0.38} & \textbf{0.13}/\textbf{0.18} & \textbf{0.35}/\textbf{0.45} & \textbf{0.43}/\textbf{0.49} & 0.20/\textbf{0.32} & 0.11/0.28 & 0.10/\textbf{0.22} & \textbf{0.10}/\textbf{0.17} & \textbf{0.25}/\textbf{0.32} & \textbf{0.20}/\textbf{0.28} & 0.06/\textbf{0.21} & \textbf{0.19}/\textbf{0.28} \\
    (109\,ms) & Async & 0.25/0.28 & 0.18/\textbf{0.22} & \textbf{0.20}/\textbf{0.29} & \textbf{0.16}/\textbf{0.21} & \textbf{0.38}/\textbf{0.46} & \textbf{0.34}/\textbf{0.42} & 0.10/0.29 & 0.08/0.24 & 0.10/\textbf{0.21} & \textbf{0.09}/\textbf{0.19} & \textbf{0.21}/\textbf{0.31} & \textbf{0.17}/\textbf{0.26} & \textbf{0.07}/\textbf{0.22} & \textbf{0.16}/\textbf{0.27} \\
     & Avg. & \textbf{0.24}/\textbf{0.27} & \textbf{0.20}/\textbf{0.25} & \textbf{0.28}/\textbf{0.33} & \textbf{0.14}/\textbf{0.20} & \textbf{0.36}/\textbf{0.46} & \textbf{0.38}/\textbf{0.45} & 0.15/\textbf{0.30} & 0.09/0.26 & 0.10/\textbf{0.21} & \textbf{0.09}/\textbf{0.18} & \textbf{0.23}/\textbf{0.31} & \textbf{0.18}/\textbf{0.27} & \textbf{0.07}/\textbf{0.21} & \textbf{0.18}/\textbf{0.27} \\
    \midrule
    X-VLA & Sync & 0.18/0.19 & 0.16/0.18 & 0.08/0.10 & 0.03/0.06 & 0.20/0.20 & 0.08/0.12 & \textbf{0.25}/0.30 & 0.18/0.27 & 0.05/0.12 & 0.03/0.09 & 0.15/0.18 & 0.09/0.15 & \textbf{0.07}/0.13 & 0.11/0.15 \\
    (156\,ms) & Async & \textbf{0.28}/\textbf{0.32} & \textbf{0.19}/0.21 & 0.05/0.08 & 0.01/0.07 & 0.28/0.29 & 0.14/0.18 & \textbf{0.23}/\textbf{0.30} & \textbf{0.19}/\textbf{0.29} & 0.05/0.14 & 0.03/0.06 & 0.18/0.22 & 0.11/0.16 & 0.06/0.14 & 0.12/0.18 \\
     & Avg. & 0.23/0.26 & 0.17/0.20 & 0.06/0.09 & 0.02/0.07 & 0.24/0.24 & 0.11/0.15 & \textbf{0.24}/\textbf{0.30} & \textbf{0.18}/\textbf{0.28} & 0.05/0.13 & 0.03/0.08 & 0.16/0.20 & 0.10/0.15 & \textbf{0.07}/0.13 & 0.11/0.16 \\
    \midrule
    FastWAM & Sync & 0.05/0.07 & 0.08/0.10 & 0.00/0.06 & 0.00/0.03 & 0.00/0.00 & 0.00/0.00 & 0.20/0.31 & \textbf{0.19}/\textbf{0.30} & \textbf{0.18}/\textbf{0.22} & 0.06/0.09 & 0.09/0.13 & 0.07/0.10 & 0.02/0.08 & 0.06/0.11 \\
    (236\,ms) & Async & 0.03/0.04 & 0.09/0.12 & 0.00/0.04 & 0.00/0.03 & 0.00/0.02 & 0.00/0.01 & 0.05/0.27 & 0.06/0.24 & \textbf{0.18}/\textbf{0.21} & 0.08/0.10 & 0.05/0.11 & 0.05/0.10 & 0.05/0.15 & 0.05/0.11 \\
     & Avg. & 0.04/0.05 & 0.08/0.10 & 0.00/0.05 & 0.00/0.03 & 0.00/0.01 & 0.00/0.00 & 0.12/0.29 & 0.12/0.27 & \textbf{0.18}/\textbf{0.21} & 0.07/0.10 & 0.07/0.12 & 0.06/0.10 & 0.04/0.11 & 0.06/0.11 \\
    \bottomrule
  \end{tabular}%
  }
\end{table*}

\section{Experiments}
\label{sec:results}

In this section, we aim to answer four research questions:
\textbf{RQ1.} How do VLAs and WAMs differ in their action capabilities in dynamic environments?
\textbf{RQ2.} What do WAMs' future predictions reveal about their physical anticipation in dynamic environments?
\textbf{RQ3.} How much does inference latency matter in solving tasks under different evaluation regimes?
\textbf{RQ4.} Do policies generalize to unseen dynamic compositions?

\paragraph{R1: the VLAs are better at reactive tracking, while FastWAM captures simple physical dynamics more effectively.}
As shown in Table~\ref{tab:results-sub}, $\pi_{0.5}$ achieves both the highest overall success rate and progress score, and leads under nearly all setups within Trajectory Prediction, Periodic Pattern, and Dynamic Avoidance. The exception is the asynchronous setting of Trajectory Prediction, where X-VLA achieves higher SR in both easy and medium settings. Comparing performance across categories, VLAs perform particularly well on tasks involving consistently moving objects, as reflected in the three categories discussed above.

\begin{figure}[t]
  \centering
  \includegraphics[width=\columnwidth]{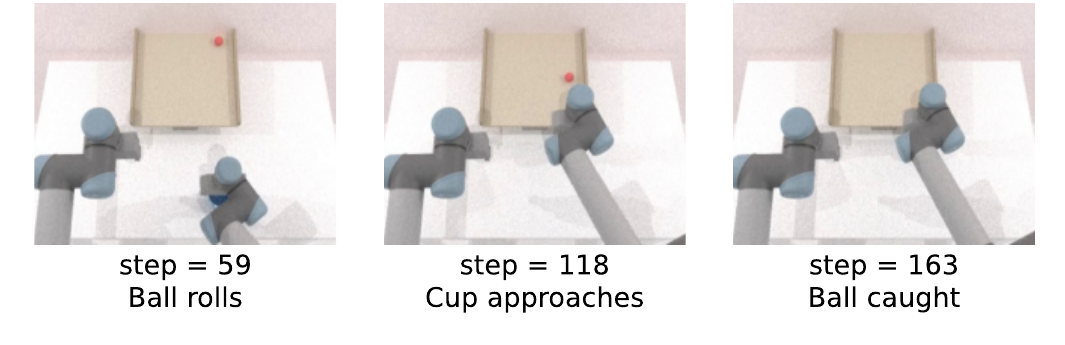}
  \caption{A successful $\pi_{0.5}$ rollout on Catch Ramp Ball under the base, synchronous protocol. }
  \label{fig:r1-policy-rollouts}\vspace{-0.3cm}
\end{figure}

A close inspection of rollout videos reveals that VLAs primarily track the moving object or wait at a location from which the task can be completed, i.e. they rely on reactive behavior and trial-and-error rather than anticipating future states (Fig.~\ref{fig:r1-policy-rollouts}). We further observe that, in distractor-heavy settings, such as Household environments, performance is lower than in the corresponding conceptual setting. This can be partially due to higher scene clutter, task semantics, or the physical phenomena. Employing such a reactive strategy $\pi_{0.5}$ reaches at most $0.43$ in any single category-difficulty-protocol and $0.19$ SR overall, leaving most episodes unsolved.

Despite its lower overall performance, FastWAM achieves the highest success rate on Static-to-Dynamic Understanding in the easy setting ($0.18$ compared with $0.10$ and $0.05$), which is the weakest category for both VLAs. On state transition tasks, where most scenes are in a static state, FastWAM is competitive with $\pi_{0.5}$ and achieves best performance on medium difficulty under synchronous protocol. These results indicate that FastWAM can solve tasks in which the dynamics are simple but must be understood before the task can be completed, demonstrating a basic ability to anticipate physical state transitions. However, this advantage decreases substantially as the dynamics become more complex. FastWAM's advantage is mainly in the easy setting: on Static-to-Dynamic Understanding both SR and PS fall by more than half from easy to medium whereas on State Transition its SR is unchanged. Averaged over all conceptual tasks the decline is generally small. This is consistent with FastWAM's predictive model degrading as the dynamics become more complex (see R2).

\paragraph{R2: FastWAM anticipates action-induced changes but struggles with pre-existing scene dynamics.} 

FastWAM ranks last, with an overall success rate of only $0.06$, despite explicitly predicting the future. To investigate the underlying reason, we compare FastWAM's predicted videos with recorded rollouts for all 30 tasks under the base (easy), synchronous protocol. FastWAM achieves 29.30\,dB PSNR and 0.778 SSIM overall. Because these metrics are computed over the full frame, they are dominated by static background pixels, hence can remain high even when a small moving object is mispredicted. In our measurements, prediction fidelity and task success are not monotonically related across categories. On periodic-pattern tasks, FastWAM achieves 30.33\,dB PSNR and 0.784 SSIM, but $0.00$ SR and  only $0.06$ PS. On state-transition tasks, FastWAM achieves lower prediction fidelity (28.69\,dB PSNR and 0.774 SSIM), yet higher task performance ($0.20$ SR and $0.31$ PS). 

\begin{figure}[t]
  \centering
  \includegraphics[width=\columnwidth]{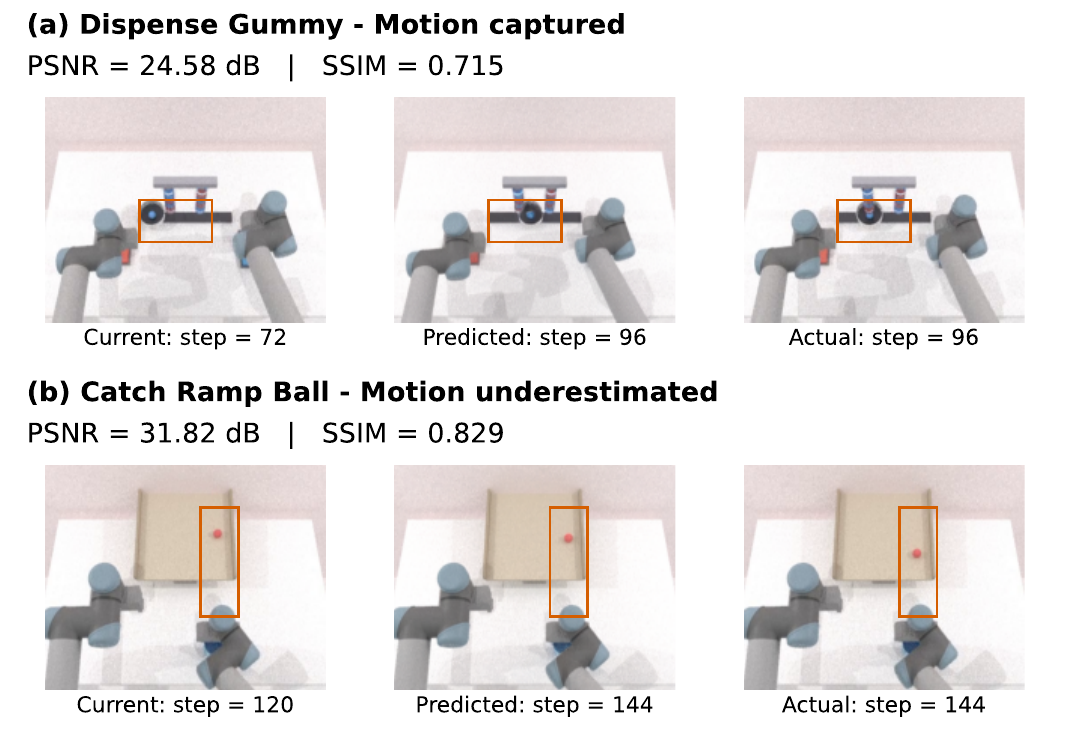}
  \caption{Comparison of FastWAM predictions with ground-truth rollouts under the base, synchronous protocol. }
  \label{fig:r2-prediction-comparison}
\end{figure}

To examine whether key dynamic events and task-completing robot actions are correctly predicted, we conduct a qualitative analysis of the recorded rollouts. The conclusion follows R1: FastWAM captures simple physical dynamics starting from static states. The predictions, however, are not uniformly accurate. For instance, as shown in Fig.~\ref{fig:r2-prediction-comparison},  FastWAM captures the displacement of the bowl due to the gripper pressing the button in a successful rollout but overestimates the motion of the bowl. When the scene is already dynamic at the start of the episode, FastWAM predictions either reproduce a near-static copy of the input frame or omit the moving object. On all four periodic-pattern tasks, the predictions are incorrect or inconsistent, which potentially led to FastWAM's zero success on both the periodic-pattern and dynamic-avoidance categories. Across all 30 tasks, we observe strong consistency between prediction and action: FastWAM never achieves a nonzero success rate on tasks where its predicted dynamics are incorrect. On tasks whose dynamics its video branch does not predict correctly, FastWAM achieves zero success.

\paragraph{R3: inference latency changes which episodes policies succeed on more than their average success rate.}

During asynchronous evaluation, we measured the mean inference delays of the policies as follows:  $109\,\mathrm{ms}$ for $\pi_{0.5}$, $156\,\mathrm{ms}$ for X-VLA, and $236\,\mathrm{ms}$ for FastWAM.
Despite such delays, the aggregate differences between synchronous and asynchronous evaluation are modest: Table~\ref{tab:results-sub} shows $\pi_{0.5}$'s SR decreases by $0.03$ and FastWAM's by $0.01$ whereas X-VLA's SR increases by $0.01$.

\begin{figure}[t]
  \centering
  \includegraphics[width=\columnwidth]{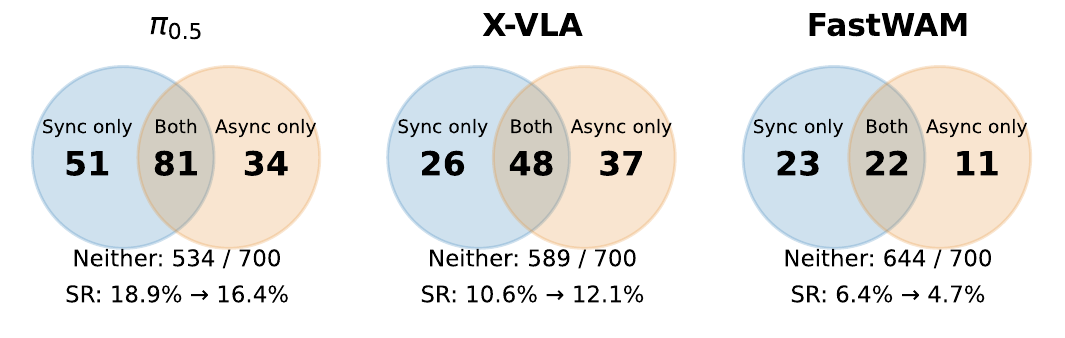}
  \caption{Overlap of successful sync and async episodes. Each policy has
  700 matched task--configuration--seed pairs, excluding held-out compositions.}
  \label{fig:r3-success-overlap}
\end{figure}

The observed small net changes conceal substantial turnover in the episodes on which each policy succeeds. Across 700 matched task--configuration--seed pairs per policy, Fig.~\ref{fig:r3-success-overlap} shows 85, 63, and 34 changed outcomes for $\pi_{0.5}$, X-VLA, and FastWAM, respectively.
Stable average performance therefore does not imply robustness to latency, but instead masks successes lost and gained between synchronous and asynchronous evaluations. The pattern is consistent with success depending on the alignment between action timing and scene dynamics: a delay can cause an action to miss an opportunity in one episode while bringing it into alignment in another.

Inference delay shows particular negative impact for two types of tasks.
First, in time-critical tasks with a finite action window, such as sorting an apple before it leaves the belt, inference delay reduces the time available to act. These tasks concentrate in the State Transition category. Second, in tasks with periodic patterns, inference delay can disrupt action timing and make it harder for policies to act in accordance with learned patterns. The largest drops are observed for $\pi_{0.5}$ on State Transition ($-0.10$) and Periodic Pattern ($-0.15$), and for FastWAM on State Transition ($-0.15$)

\begin{figure}[t]
  \centering
  \includegraphics[width=\columnwidth]{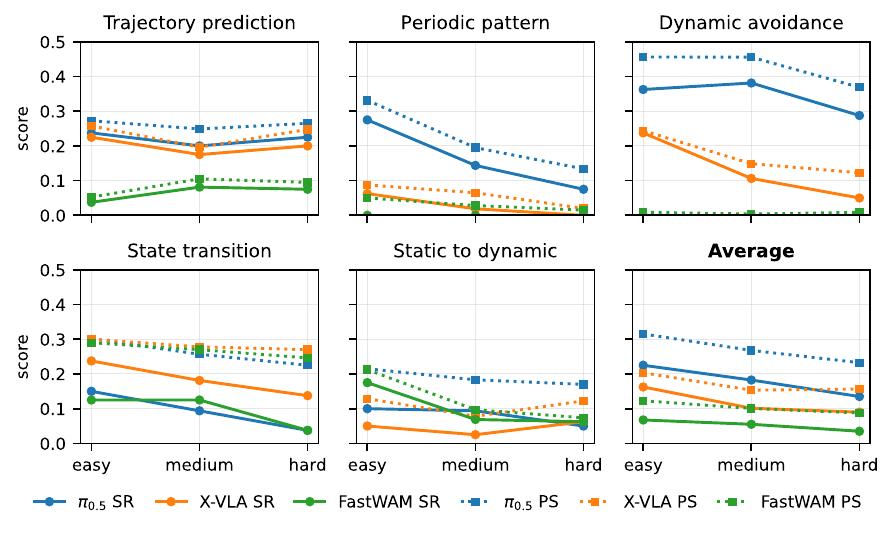}
  \caption{Success rate (solid) and progress score (dotted) across the
  difficulty ladder, by capability category and pooled over both protocols. \emph{Average} pools the five categories.}
  \label{fig:r4-lines-sub} \vspace{-0.3cm}
\end{figure}

\paragraph{R4: performance on held-out variation compositions degrades proportionally, but remains near the floor in absolute terms.} 
To test generalizability of the policies, we evaluate their performance under hard conditions (combined var1+2), which are not encountered during training. As shown in Fig.~\ref{fig:r4-lines-sub}, success rates decline under the combined variations, as expected: $\pi_{0.5}$ achieves $0.23/0.18/0.14$, X-VLA achieves $0.16/0.10/0.09$, and FastWAM achieves $0.07/0.06/0.04$ across easy, medium, and hard settings, respectively, retaining $56\%$--$61\%$ of their easy-setting performance. For both VLAs the relative decline from medium$\to$hard is no larger than from easy$\to$medium, whereas the two declines are similar for FastWAM. This shows composition adds no new source of difficulty beyond its constituent variations. 

Performance declines are most pronounced on \emph{periodic-pattern} tasks, the only category in which both VLAs degrade monotonically with increasing difficulty. Most periodic-pattern variations either introduce distractor objects that the policy must ignore (e.g., a distractor marble, decoy hole, or distractor rabbit) or remove useful affordances, such as retries or clear views, both of which appear challenging for the policies. By contrast, policies show the smallest performance declines on \emph{trajectory-prediction} tasks and even improve in some settings. Three of the four tasks under this category share a similar variation design, combining a wall-bounce variation with a distractor ball while the target ball moves across a surface. The underlying dynamics are repeatedly encountered during training, which may explain the relatively small performance degradation. 

\section{Human Experiments} 
For human experiments, we recruited twelve participants \footnote{The study was determined to be exempt from formal IRB review. Participants voluntarily agreed to participate and gave explicit consent for anonymized data collection.}, each with a background in robotics and varying degrees of familiarity with robot simulators, 3D interaction software, and video games. Each participant was assigned one conceptual task from each category and two to three household tasks, assigned round-robin so that each task received an equal number of participants. The participants completed their assigned tasks under both teleoperation and scene-interactive control. 

 For each control regime and task set, there was a series of tutorial scenarios that the participants could attempt in order to familiarize themselves with the controls and basic task mechanics. For each conceptual task, participants receive one warm-up attempt on each of \texttt{base}, \texttt{var1}, and \texttt{var2}, followed by five scored attempts on the hard \texttt{var1+2} composition. For each household task, participants receive two warm-up attempts followed by five scored attempts.
\begin{figure}[t]
  \centering
  \includegraphics[width=\columnwidth]{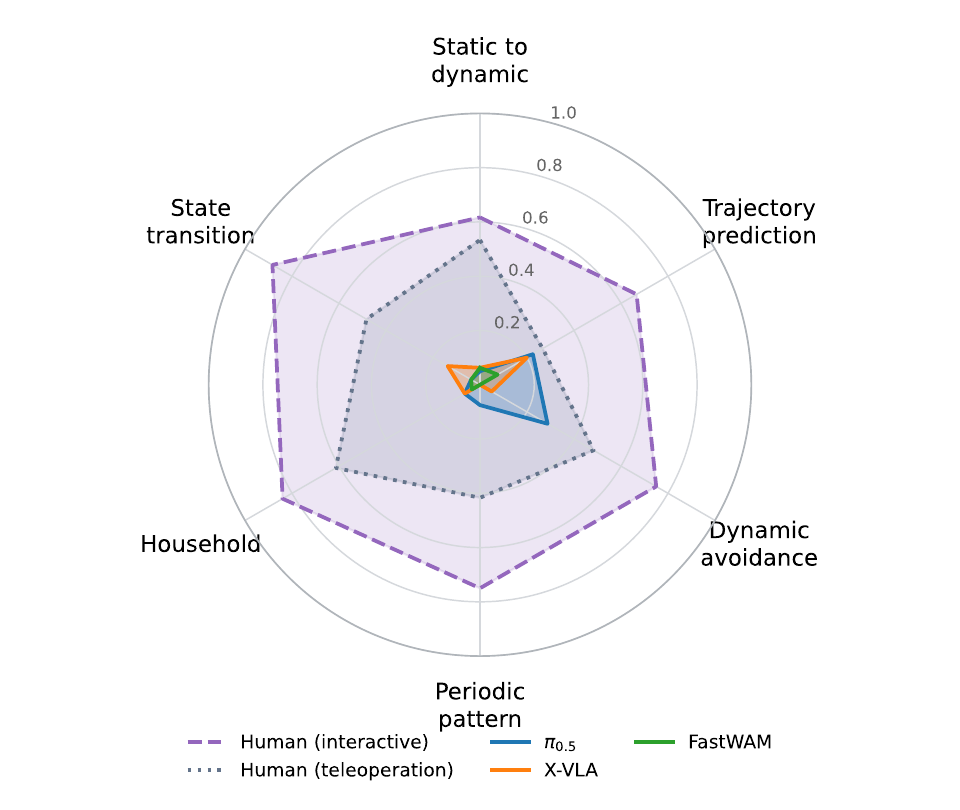}
  \caption{Success rate by category on the human-evaluated slice (conceptual tasks at \texttt{var1+2}, household tasks at \texttt{base}). }\vspace{-0.3cm}
  \label{fig:human-bars}
\end{figure}
Fig.~\ref{fig:human-bars} compares the twelve
participants' performance with the policies at the unseen \texttt{var1+2}
condition. Under teleoperation, humans solved $50.2\%$ of episodes compared with $14\%$ for the best policy on the same conditions. Under interactive control, which removes low-level control (hence not directly comparable to policies), the humans solved $77.1\%$ of the episodes. Acting through the arms consumes time that scene-interactive control does not, which is why the gap between regimes is largest on time-critical tasks: human success rates fall from $0.67$ to $0.27$ on trajectory-prediction tasks and from $0.75$ to $0.42$ on periodic-pattern tasks when switching from interactive control to teleoperation. Humans showed the smallest decrease on static-to-dynamic tasks ($0.62$ to $0.53$).

The human baseline also provides a reference for the difficulty of the tasks themselves. Under interactive control, with the control burden removed, the hardest categories are static-to-dynamic understanding and trajectory prediction, where the prediction itself is naturally harder (a ball moving on a ramp, marbles moving through a maze), while humans achieve their highest success rates on everyday state-transition tasks ($0.88$) and household tasks ($0.84$). The VLAs are relatively strongest where humans struggle most: on this slice $\pi_{0.5}$ peaks on dynamic avoidance ($0.29$) and X-VLA on trajectory prediction ($0.20$), although $\pi_{0.5}$ reaches $0.23$, categories where tracking a visible object and reacting suffices; both collapse on periodic pattern ($0.07$ and $0.00$), which humans handle at $0.75$. The per-category values quoted here are the ones plotted in Fig.~\ref{fig:human-bars}. The VLAs' behavior is consistent with a largely reactive observation-to-action mapping, whereas the participants remained successful on periodic-pattern tasks where both VLAs were near zero. FastWAM is the weakest on the categories requiring anticipation of independent motion, as discussed in R2, reaching zero on dynamic avoidance and periodic pattern.

\section{Conclusion}
We proposed \RoboDyna{}, a benchmark that jointly evaluates physical prediction and action across five capability categories with composable difficulty factors, paired synchronous and asynchronous  evaluation protocols, and interactive tools for establishing human baselines. The evaluated VLAs and the WAM showed distinct performance profiles, but all three solved fewer than one in five episodes overall. The WAM leads on easy static-to-dynamic tasks but ranks last overall. Its predictions deteriorate when objects move independently, coinciding with failures in action execution. We observed that average performance changes only modestly across evaluation protocols but the outcomes of a small percentage of episodes flip, so aggregate success rate is a poor indicator of latency robustness. We further showed that policies retain most of their success on compositions of
dynamics left out of training, and that their overall performance remains well below humans.

\bibliographystyle{IEEEtran}
\bibliography{references}

\end{document}